\documentclass{article}

\usepackage[final, nonatbib]{nips_2018}

\usepackage{subfig}
\usepackage{multicol}
\usepackage{enumitem}

\usepackage[utf8]{inputenc} % allow utf-8 input
\usepackage[T1]{fontenc}    % use 8-bit T1 fonts
\usepackage{hyperref}       % hyperlinks
\usepackage{url}            % simple URL typesetting
\usepackage{booktabs}       % professional-quality tables
\usepackage{amsfonts}       % blackboard math symbols
\usepackage{amsmath}
\usepackage{nicefrac}       % compact symbols for 1/2, etc.
\usepackage{microtype}      % microtypography
\usepackage{graphicx}

\newcommand{\comment}[1]{}

\title{Data-Efficient Hierarchical Reinforcement Learning}

\author{
  Ofir Nachum \\
  Google Brain\\
  \texttt{ofirnachum@google.com} \\
  \And
  Shixiang Gu\thanks{Also at University of Cambridge; Max Planck Institute of Intelligent Systems.} \\
  Google Brain \\
  \texttt{shanegu@google.com} \\
  \AND
  Honglak Lee \\
  Google Brain \\
  \texttt{honglak@google.com} \\
  \And
  Sergey Levine\thanks{Also at UC Berkeley.} \\
  Google Brain \\
  \texttt{slevine@google.com} \\
}

\begin{document}

\maketitle

\setcounter{footnote}{0}

\begin{abstract}
  Hierarchical reinforcement learning (HRL) is a promising approach to extend traditional reinforcement learning (RL) methods to solve more complex tasks.
  Yet, the majority of current HRL methods require careful task-specific design and on-policy training, making them difficult to apply in real-world scenarios.
  In this paper, we study how we can develop HRL algorithms that are general, in that they do not make onerous additional assumptions beyond standard RL algorithms, and efficient, in the sense that they can be used with modest numbers of interaction samples, making them suitable for real-world problems such as robotic control. 
  For generality, 
  we develop a scheme where lower-level controllers are supervised with goals that are learned and proposed automatically by the higher-level controllers.
  %The lower-level policy is trained to match its observations to these goals.
  %In this way, the lower-level policy is trained to perform basic locomotion, while the higher-level policy is able to focus its planning and (crucially) its explorative behavior on the main task.
  To address efficiency, we propose to use off-policy experience for both higher- and lower-level training. This poses a considerable challenge, since changes to the lower-level behaviors change
  the action space
  for the higher-level policy, and we introduce an off-policy correction to remedy this challenge.
This allows us to take advantage of recent advances in off-policy model-free RL to learn both higher- and lower-level policies using substantially fewer environment interactions than on-policy algorithms.
We term the resulting HRL agent {\em HIRO} and find that it is generally applicable and highly sample-efficient.
  Our experiments show that HIRO can be used to learn highly complex behaviors for simulated robots, such as pushing objects and utilizing them to reach target locations,\footnote{See videos at \url{https://sites.google.com/view/efficient-hrl}} learning from only a few million samples, equivalent to a few days of real-time interaction. 
  In comparisons with a number of prior HRL methods, we find that our approach substantially outperforms previous state-of-the-art techniques.\footnote{Find open-source code at \url{https://github.com/tensorflow/models/tree/master/research/efficient-hrl}}
\end{abstract}

\section{Introduction}
Deep reinforcement learning (RL) has made significant progress on a range of continuous control tasks, such as locomotion skills~\cite{schulman2015trust,ddpg,heess2017emergence}, %simulated driving~\cite{ddpg}, 
learning dexterous manipulation behaviors~\cite{rajeswaran}, and training robot arms for simple manipulation tasks~\cite{gu2017deep,vevcerik2017leveraging}.
However, most of these behaviors are inherently atomic: they require performing some simple skill, either episodically or cyclically, and rarely involve complex multi-level reasoning, such as utilizing a variety of locomotion behaviors to accomplish complex goals that require movement, object interaction, and discrete decision-making. 

Hierarchical reinforcement learning (HRL), in which multiple layers of policies are trained to perform decision-making and control at successively higher levels of temporal and behavioral abstraction, has long held the promise to learn such difficult tasks~\cite{dayan1993feudal,parr1998reinforcement,sutton1999between,barto2003recent}.
By having a hierarchy of policies, of which only the lowest applies actions to the environment, one is able to train the higher levels to plan over a longer time scale.  Moreover, if the high-level actions correspond to semantically different low-level behavior, standard exploration techniques may be applied to more appropriately explore a complex environment.
Still, there is a large gap between the basic definition of HRL and the promise it holds to successfully solve complex environments.
%In the text of this paper, we will focus on two-layer hierarchies, and thus only refer to a {\em lower-level} and a {\em higher-level} policy.
To achieve the benefits of HRL, there are a number of questions that one must suitably answer:
How should one train the lower-level policy to induce semantically distinct behavior?  How should the high-level policy actions be defined?  How should the multiple policies be trained without incurring an inordinate amount of experience collection?  Previous work has attempted to answer these questions in a variety of ways and has provided encouraging successes~\cite{vezhnevets2017feudal,florensa2017stochastic,frans2017meta,heess2016learning,sigaud2018policy}.
However, many of these methods lack generality, requiring some degree of manual task-specific design, and often require expensive on-policy training that is unable to benefit from advances in off-policy model-free RL, which in recent years has drastically brought down sample complexity requirements~\cite{td3,sac,barth2018distributed}.

\begin{figure}
  \begin{center}
  \addtolength{\tabcolsep}{-3pt}
  	\begin{tabular}{cccc}
    \includegraphics[width=0.13\textwidth]{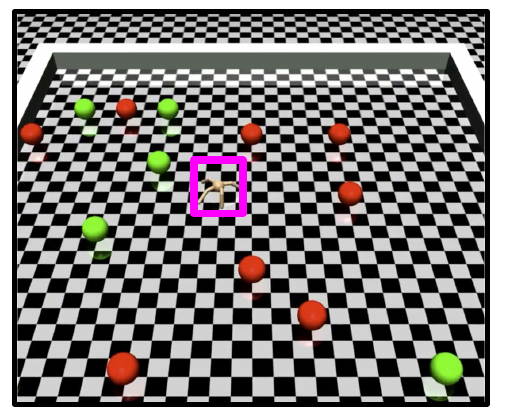} &
    \includegraphics[width=0.26\textwidth]{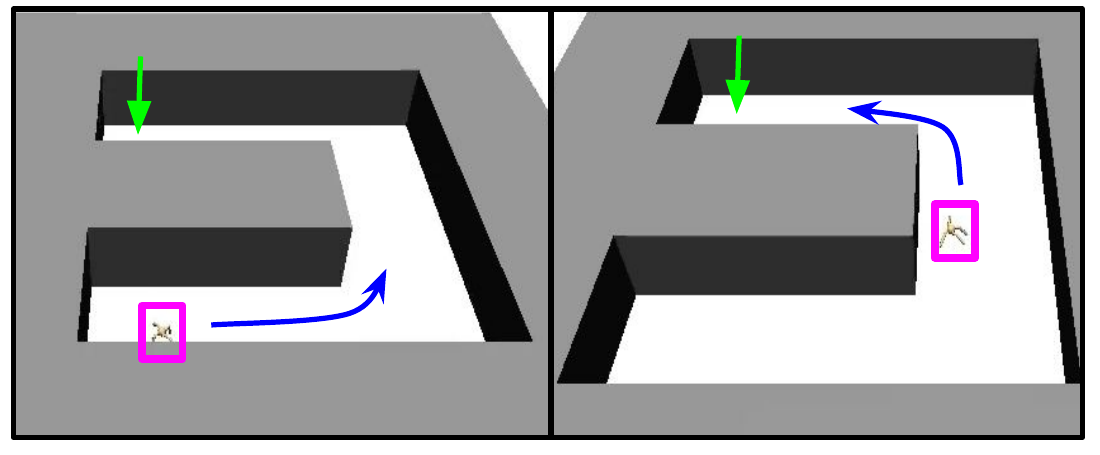} &
    \includegraphics[width=0.28\textwidth]{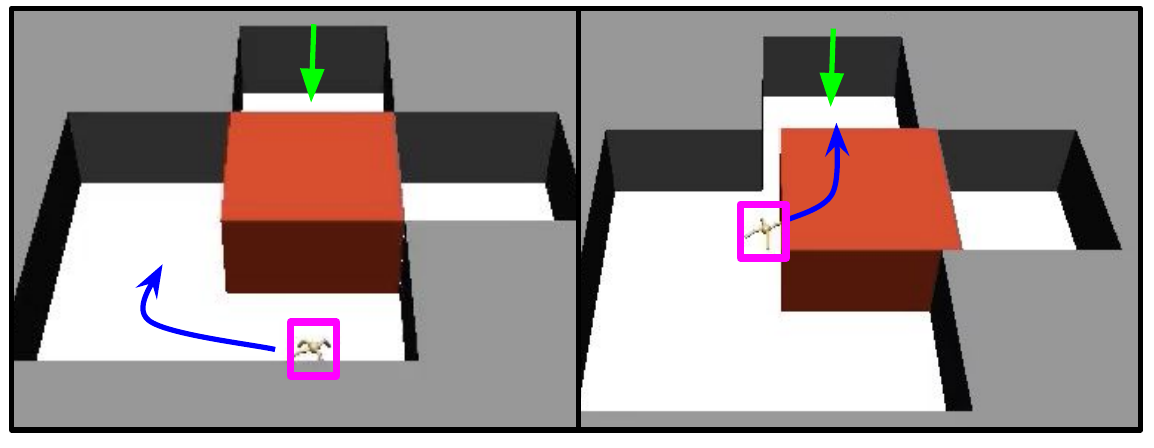} &
    \includegraphics[width=0.255\textwidth]{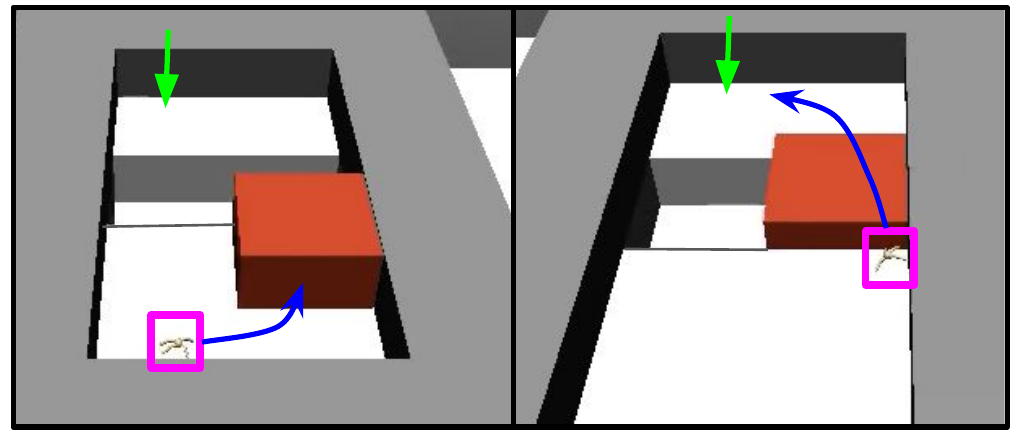} %\\
    %\multicolumn{3}{c}{\includegraphics[width=0.4\textwidth]{ant_fall_b}}
    \end{tabular}
    \addtolength{\tabcolsep}{3pt}
  \end{center}
  \caption{The Ant Gather task along with the three hierarchical navigation tasks we consider: Ant Maze, Ant Push, and Ant Fall.  The ant (magenta rectangle) is rewarded for approaching the target location (green arrow). A successful policy must perform a complex sequence of directional movement and, in some cases, interact with objects in its environment (red blocks); e.g., pushing aside an obstacle (second from right) or using a block as a bridge (right). In our HRL method, a higher-level policy periodically produces goal states (corresponding to desired positions and orientations of the ant and its limbs), which the lower-level policy is rewarded to match (blue arrow).}
	\label{fig:envs}
\end{figure}

%We propose instead to 
For generality, we propose to
take advantage of the state observation provided by the environment to the agent, which in locomotion tasks can include the position and orientation of the agent and its limbs.
We let the high-level actions be goal states and reward the lower-level policy for performing actions which yield it an observation close to matching the desired goal.
In this way, our HRL setup does not require a manual or multi-task design and is fully general.

This idea of a higher-level policy commanding a lower-level policy to match observations to a goal state has been proposed before~\cite{dayan1993feudal,vezhnevets2017feudal}.
Unlike previous work, which represented goals and rewarded matching observations within a learned embedding space, we use the state observations in their raw form.
This significantly simplifies the learning, and in our experiments, we observe substantial benefits for this simpler approach.

While these goal-proposing methods are very general, they require training with on-policy RL algorithms, which are generally less efficient than off-policy methods~\cite{gu2016q,tpcl}. 
On-policy training has been attractive in the past since, outside of discrete control, off-policy methods have been plagued with instability~\cite{gu2016q}, which is amplified when training multiple policies jointly, as in HRL.
Other than instability, off-policy training poses another challenge that is unique to HRL. Since the lower-level policy is changing underneath the higher-level policy, a sample observed for a certain high-level action in the past may not yield the same low-level behavior in the future, and thus not be a valid experience for training. This amounts to a non-stationary problem for the higher-level policy. We remedy this issue by introducing an off-policy correction, which re-labels an experience in the past with a high-level action chosen to maximize the probability of the past lower-level actions.  In this way, we are able to use past experience for training the higher-level policy, taking advantage of progress made in recent years to provide stable, robust, and general off-policy RL methods~\cite{td3, tpcl, barth2018distributed}.

In summary, we introduce a method to train a multi-level HRL agent that stands out from previous methods by being both generally applicable and data-efficient.  Our method achieves generality by training the lower-level policy to reach goal states learned and instructed by the higher-levels.
In contrast to prior work that operates in this goal-setting model, we use states as goals directly, which allows for simple and fast training of the lower layer. % and achieves substantially better performance on a range of challenging continuous control tasks.
Moreover, by using off-policy training with our novel off-policy correction, our method is extremely sample-efficient.
We evaluate our method on several difficult environments.  
These environments require the ability to perform exploratory navigation as well as complex sequences of interaction with objects in the environment (see Figure~\ref{fig:envs}). 
While these tasks are unsolvable by existing non-HRL methods,
we find that our HRL setup can learn successful policies.
When compared to other published HRL methods, we also observe the superiority of our method, in terms of both final performance and speed of learning.  In only a few million experience samples, our agents are able to adequately solve previously unapproachable tasks.

\section{Background}
We adopt the standard continuous control RL setting, in which an agent interacts with an environment over periods of time according to a behavior policy $\mu$.  At each time step $t$, the environment produces a state observation $s_t\in\mathbb{R}^{d_s}$.%, where $d_s$ is the dimensionality of the states.  
The agent then samples an action $a_t\sim\mu(s_t),a_t\in\mathbb{R}^{d_a}$%, where $d_a$ is the action dimension, 
and applies the action to the environment.  
The environment then yields a reward $R_t$ sampled from an unknown reward function $R(s_t, a_t)$
and either terminates the episode at state $s_T$ or transitions to a new state $s_{t+1}$ sampled from an unknown transition function $f(s_t, a_t)$.
The agent's goal is to maximize the expected future discounted reward %\begin{equation}
$\mathbb{E}_{s_{0:T}, a_{0:T-1}, R_{0:T-1}}\left[\textstyle\sum_{i=0}^{T-1} \gamma^i R_i\right]$,
%\end{equation}
where $0\le\gamma<1$ is a user-specified discount factor.
A well-performing RL algorithm will learn a good behavior policy $\mu$ from (ideally a small number of) interactions with the environment.  

\subsection{Off-Policy Temporal Difference Learning}

Temporal difference learning is a powerful paradigm in RL, in which a policy may be learned efficiently from state-action-reward transition tuples $(s_t, a_t, R_t, s_{t+1})$ collected from interactions with the environment.  In our HRL method, we utilize the TD3 learning algorithm~\cite{td3}, a variant of the popular DDPG algorithm for continuous control~\cite{ddpg}.

In DDPG, a deterministic neural network policy $\mu_\phi$ is learned along with its corresponding state-action Q-function $Q_\theta$ by performing gradient updates on parameter sets $\phi$ and $\theta$.  The Q-function represents the future value of taking a specific action $a_t$ starting from a state $s_t$.  Accordingly, it is trained to minimize the average Bellman error over all sampled transitions, which is given by
\begin{equation}
\mathcal E(s_t, a_t, s_{t+1}) = (Q_\theta(s_t, a_t) - R_t - \gamma Q_\theta(s_{t+1}, \mu_\phi(s_{t+1})))^2.
\label{eq:bellman}
\end{equation}
The policy is then trained to yield actions which maximize the Q-value at each state.  That is, $\mu_\phi$ is trained to maximize $Q_\theta(s_t, \mu_\phi(s_t))$ over all $s_t$ collected from interactions with the environment.

We note that although DDPG trains a deterministic policy $\mu_\phi$, its behavior policy, which is used to collect experience during training is augmented with Gaussian (or Ornstein-Uhlenbeck) noise~\cite{ddpg}.  Therefore, actions are collected as $a_t\sim N(\mu_\phi(s_t), \sigma)$ for fixed standard deviation $\sigma$, which we will shorten as \mbox{$a_t\sim \mu_\phi(s_t)$}.  We will take advantage of the fact that the behavior policy is stochastic for the off-policy correction in our HRL method. TD3~\cite{td3} makes several modifications to DDPG's learning algorithm to yield a more robust and stable procedure.  Its main modification is using an ensemble over Q-value models and adding noise to the policy when computing the target value in Equation~\ref{eq:bellman}.

\section{General and Efficient Hierarchical Reinforcement Learning}

In this section, we present our framework for learning hierarchical policies, HIRO: HIerarchical Reinforcement learning with Off-policy correction. 
We make use of parameterized reward functions to specify a potentially infinite set of lower-level policies, 
each of which is trained to match its observed states $s_t$ to a desired goal. % specified by a higher-level policy.
The higher-level policy chooses these goals for temporally extended periods, and uses an off-policy correction
to enable it to use past experience collected from previous, different instantiations of the lower-level policy.

\subsection{Hierarchy of Two Policies}

\comment{
\begin{figure}
  \begin{center}
    \includegraphics[width=0.6\textwidth]{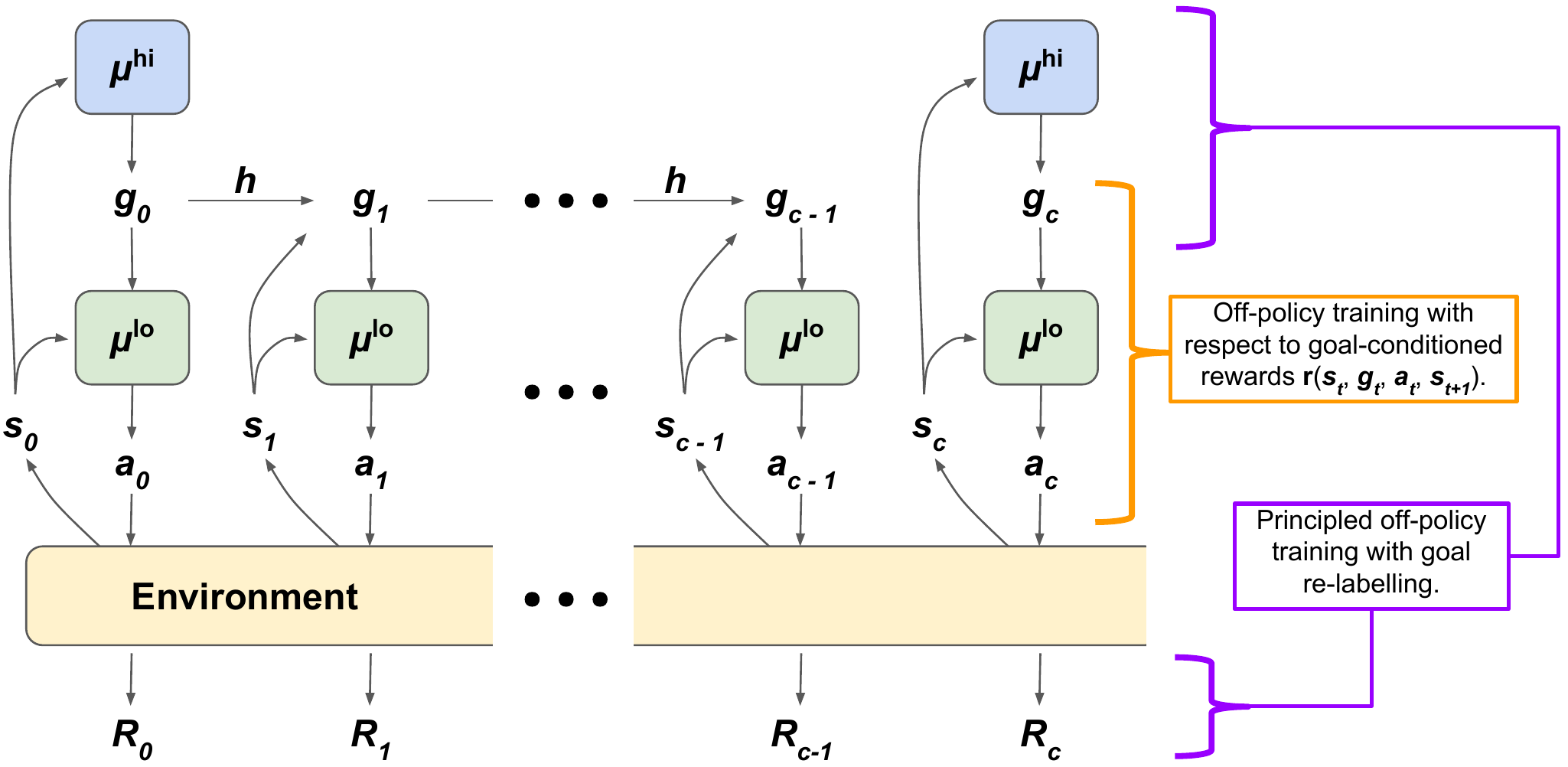}
  \end{center}
  \caption{The design of HIRO.  The lower-level policy interacts directly with the environment.  The higher-level policy instructs the lower-level policy via high-level actions, or goals, $g_t\in\mathbb{R}^{d_s}$ which it samples anew every $c$ steps. On intermediate steps, a fixed goal transition function $h$ determines the goal for the next time step. The goal simply instructs the lower-level policy to reach specific states, which allows the lower-level policy to easily learn from prior off-policy experience.}
	\label{fig:design}
\end{figure}
}

\begin{figure}
  \begin{center}
    \begin{multicols}{2}
      \begin{tabular}{ p{0.58\textwidth}   p{0.4\textwidth}}
        \includegraphics[width=0.59\textwidth]{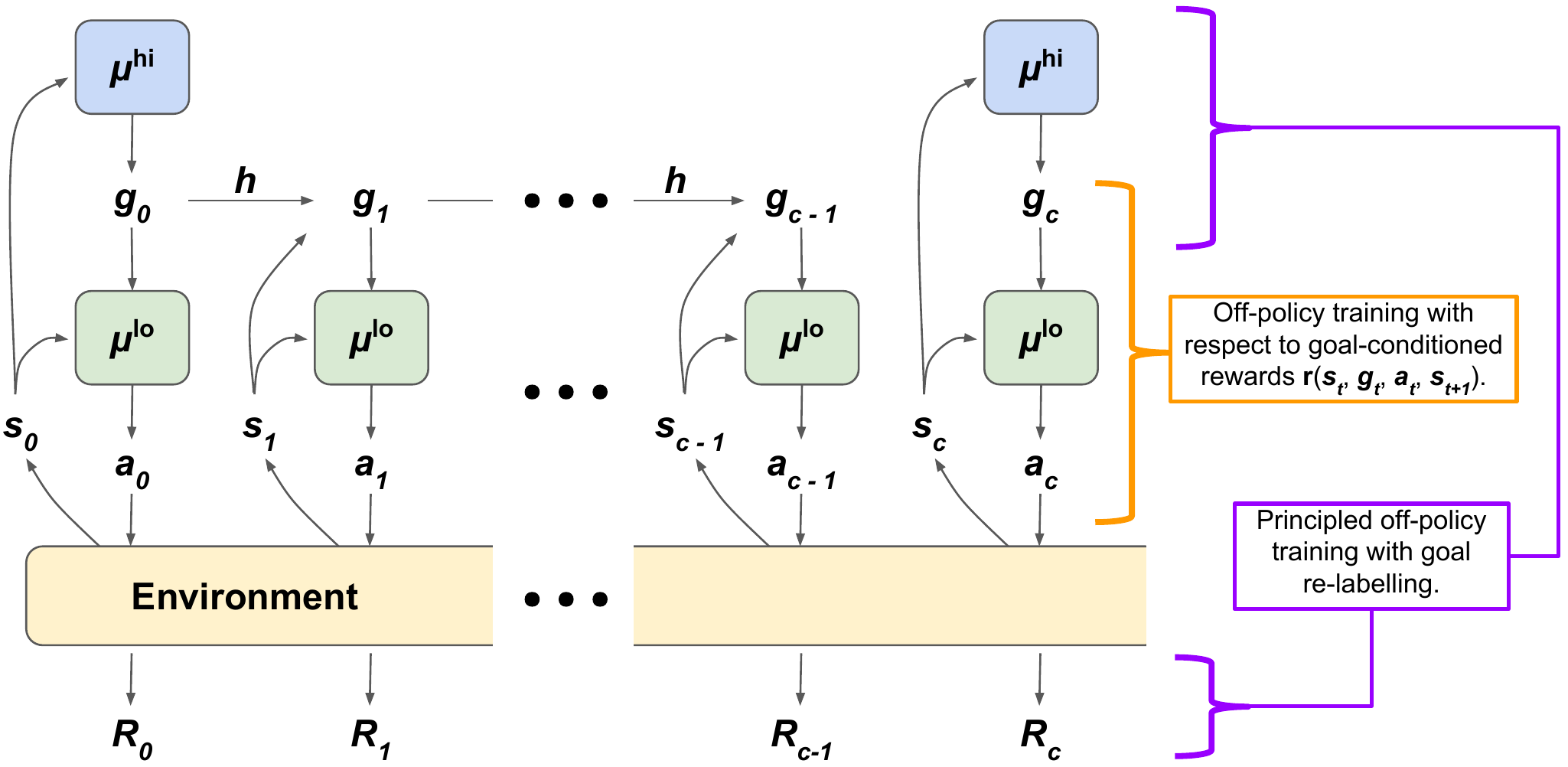}   & 
        \begin{minipage}{0.37\textwidth}
          \vspace{-1.5in}
          \begin{enumerate}[itemsep=1.1ex,leftmargin=0.1in]
            \scriptsize
            \item Collect experience $s_t$, $g_t$, $a_t$, $R_t,\dots$.
            \item Train $\mu^{lo}$ with experience transitions $(s_t, g_t, a_t, r_t, s_{t+1}, g_{t+1})$ using $g_t$ as additional state observation and reward given by goal-conditioned function $r_t=$ 
            
            $r(s_t, g_t, a_t, s_{t+1})=-||s_t+g_t-s_{t+1}||_2$.
            \item Train $\mu^{hi}$ on temporally-extended experience $(s_t,\tilde{g}_t, \sum R_{t:t+c-1}, s_{t+c})$, where $\tilde{g}_t$ is re-labelled high-level action to maximize probability of past low-level actions $a_{t:t+c-1}$.
            \item Repeat.
          \end{enumerate}
          \vspace{-0.1in}
        \end{minipage}
      \end{tabular}
    \end{multicols}
    \vspace{-0.23in}
  \end{center}
  \caption{The design and basic training of HIRO.  The lower-level policy interacts directly with the environment.  The higher-level policy instructs the lower-level policy via high-level actions, or goals, $g_t\in\mathbb{R}^{d_s}$ which it samples anew every $c$ steps. On intermediate steps, a fixed goal transition function $h$ determines the next step's goal. The goal simply instructs the lower-level policy to reach specific states, which allows the lower-level policy to easily learn from prior off-policy experience.}
  \label{fig:design}
\end{figure}

We extend the standard RL setup to a hierarchical two-layer structure,
with a lower-level policy $\mu^{lo}$ and a higher-level policy $\mu^{hi}$ (see Figure~\ref{fig:design}).
The higher-level policy operates at a coarser layer of abstraction and sets goals to the lower-level policy, which correspond directly to states that the lower-level policy attempts to reach.
At each time step $t$, the environment provides an observation state $s_t$.
The higher-level policy observes the state and produces a {\em high-level action} (or {\em goal})
$g_t\in\mathbb{R}^{d_s}$ by either sampling from its policy
$g_t\sim\mu^{hi}$ when $t \equiv 0 \pmod{c}$, 
or otherwise using a fixed goal transition function $g_t = h(s_{t-1}, g_{t-1}, s_t)$ (which in the simplest case can be a pass-through function, although we will consider a slight variation in our specific design). 
This provides temporal abstraction, since high-level decisions via $\mu^{hi}$ are made only every $c$ steps.
The lower-level policy $\mu^{lo}$ observes the state $s_t$ and goal $g_t$ and 
produces a low-level atomic action $a_t\sim\mu^{lo}(s_t, g_t)$, which is applied to the environment.
The environment then yields a reward $R_t$ sampled from an unknown reward function $R(s_t, a_t)$
and transitions to a new state $s_{t+1}$ sampled from an unknown transition function $f(s_t, a_t)$.

The higher-level controller provides the lower-level with an intrinsic reward $r_t=r(s_t, g_t, a_t, s_{t+1})$,
using a fixed parameterized reward function $r$.
The lower-level policy will store the experience $(s_t, g_t, a_t, r_t, s_{t+1}, h(s_t, g_t, s_{t+1}))$
for off-policy training.
The higher-level policy collects the environment rewards $R_t$ and, every $c$ time steps,
stores the higher-level transition $(s_{t:t+c-1}, g_{t:t+c-1}, a_{t:t+c-1}, R_{t:t+c-1}, s_{t+c})$
for off-policy training.

% Candidate for deletion:
%While our HRL setup uses a newly-generated high-level action every $c$ time steps for fixed $c$, our method is general and can easily be extended to other protocols.  For example, in some cases it may be more appropriate to use a learned or fixed lower-level termination function to determine when to re-sample a high-level action.

\subsection{Parameterized Rewards}
\begin{figure}
  \begin{center}
    \includegraphics[width=0.75\textwidth]{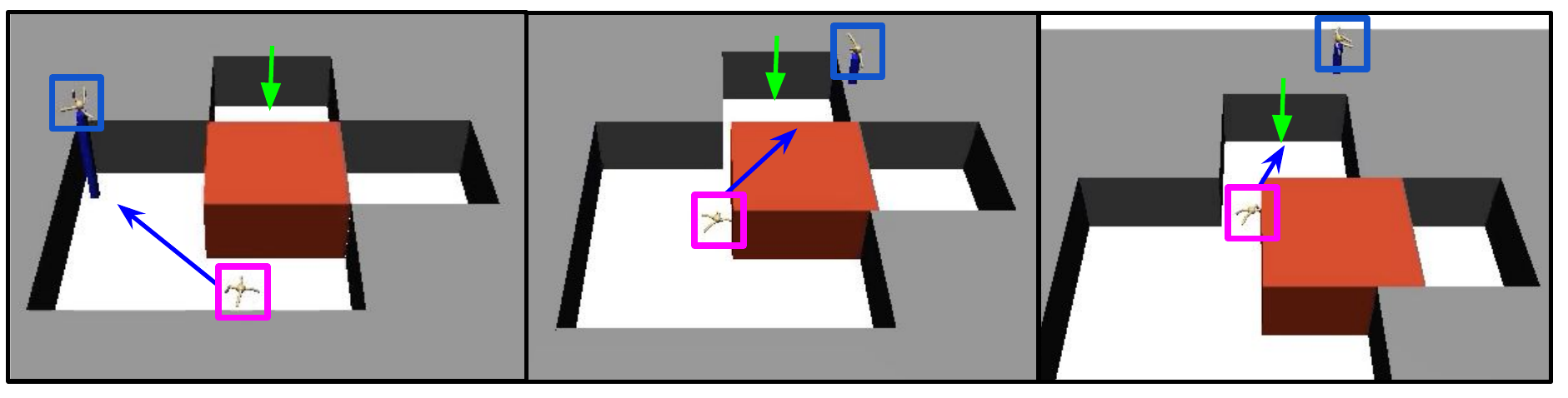}
  \end{center}
  \caption{An example of a higher-level policy producing goals in terms of desired observations, which in this task correspond to positions and orientations of all of the joints of a quadrupedal robot (including root position).  The lower-level policy has direct control of the agent (pink), and is rewarded for matching the position and orientation of its torso and each limb to the goal (blue rectangle, raised for visibility).  In this way, the two-layer policy can perform a complex task involving a sequence of movements and interactions; e.g. pushing a block aside to reach a target (green).}
	\label{fig:example}
\end{figure}

Our higher-level policy produces goals $g_t$ indicating desired relative changes in state observations.
That is, at step $t$, the higher-level policy produces a goal $g_t$, indicating its desire for the lower-level agent
to take actions that yield it an observation $s_{t+c}$ that is close to $s_t + g_t$. Although some state dimensions (e.g., the position of the quadrupedal robot in Figure~\ref{fig:example}) are more natural as goal subspaces, we chose this more generic goal representation to make it broadly applicable, without any manual design of goal spaces, primitives, or controllable dimensions. This makes our method general and easy to apply to new problem settings.
To maintain the same absolute position of the goal regardless of state change, the goal transition model $h$ is defined as
\begin{equation}
h(s_t, g_t, s_{t+1}) = s_t + g_t - s_{t+1}.
\end{equation}
We define the intrinsic reward as a parameterized reward function based on the distance between the current observation and the goal observation:
\begin{align}
r(s_t,g_t,a_t,s_{t+1}) = -||s_t + g_t - s_{t+1}||_2.
\end{align}
This rewards the lower-level policy for taking actions that yield observations that are close to the desired value $s_t + g_t$.
In our evaluations on simulated ant locomotion, we use all positional observations as the representation for $g_t$, without distinguishing between the $(x,y,z)$ root position or the joints, making for a generic and broadly applicable choice of goal space. The reward $r$ and transition function $h$ are computed only with respect to these positional observations.  See Figure~\ref{fig:example} for an example of the goals $g_t$ chosen during a successful navigation of a complex environment.

The lower-level policy may be trained using standard methods by simply incorporating $g_t$ as an additional input into the value and policy models.
For example, in DDPG, the equivalent objective to Equation~\ref{eq:bellman} in terms of lower-level Q-value function $Q_\theta^{lo}$ is to minimize the error
\begin{equation}
(Q_\theta^{lo}(s_t, g_t, a_t) - r(s_t, g_t, a_t, s_{t+1}) - \gamma Q_\theta^{lo}(s_{t+1}, g_{t+1}, \mu_\phi^{lo}(s_{t+1}, g_{t+1})))^2,
\end{equation}
for all transitions $(s_t, g_t, a_t, s_{t+1}, g_{t+1})$.  The policy $\mu_\phi^{lo}$ would be trained to maximize the Q-value $Q_\theta^{lo}(s_t, g_t, \mu_\phi^{lo}(s_t, g_t))$
for all sampled state-goal tuples $(s_t, g_t)$.

Parameterized rewards are not a new concept, and have been studied previously~\cite{uvf,held2017automatic}.
They are a natural choice for a generally applicable HRL method and have therefore appeared 
as components of other HRL methods~\cite{vezhnevets2017feudal,kulkarni2016hierarchical,plappert2018multi,levy2017hierarchical}. A significant distinction between our method and these prior approaches is that we directly use the state observation as the goal, and changes in the state observation as the action space for the higher-level policy, in contrast to prior methods that must train the goal representation. This allows the lower-level policy to begin receiving reward signals immediately, even before the lower-level policy has figured out how to reach the goal and before the task's extrinsic reward provides any meaningful supervision. In our experiments (Section~\ref{sec:experiments}), we find that this produces substantially better results.

\subsection{Off-Policy Corrections for Higher-Level Training}
While a number of prior works have proposed two-level HRL architectures that involve some sort of goal setting, such designs in previous work generally require on-policy training~\cite{vezhnevets2017feudal}. This is because the changing behavior of the lower-level policy creates a non-stationary problem for the higher-level policy, and old off-policy experience may exhibit different transitions conditioned on the same goals. However, for HRL methods to be applicable to real-world settings, they must be sample-efficient, and off-policy algorithms (often based on some variant of Q-function learning) generally exhibit substantially better sample efficiency than on-policy actor-critic or policy gradient variants. In this section, we describe how we address the challenge of off-policy training of the higher-level policy.

We would like to take the higher-level transition tuples $(s_{t:t+c-1}, g_{t:t+c-1}, a_{t:t+c-1}, R_{t:t+c-1}, s_{t+c})$, where $x_{t:t+c-1}$ denotes the sequence $x_t,\dots,x_{t+c-1}$,
which are collected by the higher-level policy and convert them to state-action-reward transitions
$(s_t, g_t, \sum R_{t:t+c-1}, s_{t+c})$ that can be pushed into the replay buffer of any standard off-policy RL algorithm.
However, since transitions obtained from past lower-level controllers do not accurately reflect the actions (and therefore resultant states $s_{t+1:t+c}$) that would occur if the same goal were used with the current lower-level controller, we must introduce a correction that translates old transitions into ones that agree with the current lower-level controller.% each time that controller is updated.

%The high-level state transition from $s_t,g_t$ to $R_{t:t+c},s_{t+c}$ induced by a past instantiation of the lower-level policy is not necessarily the behavior that would be induced by the same policy at a later point in training.
%For this reason, extra care is necessary when performing off-policy training.

%Unfortunately, the low-level policy used for a past high-level transition $(s_t, g_t, \sum R_{t:t+c}, s_{t+c})$ may not be easily changed to determine the correct transition $s_t,g_t$ to $\tilde{R}_{t:t+c},\tilde{s}_{t+c}$, since changing it would require re-deriving the low-level state transitions $s_{t:t+c}$ and rewards $R_{t:t+c}$, which are determined by unknown environment dynamics.
Our main observation is that the goal $g_t$ of a past high-level transition $(s_t, g_t, \sum R_{t:t+c-1}, s_{t+c})$ may be changed to make the actual observed action sequence more likely to have happened with respect to the current instantiation of $\mu^{lo}$.
The high-level action $g_t$ which in the past induced a low-level behavior $a_{t:t+c-1}\sim\mu^{lo}(s_{t:t+c-1}, g_{t:t+c-1})$ may be re-labeled to a goal $\tilde{g}_t$ which is likely to induce the same low-level behavior with the current instantiation of the lower-level policy.
Thus, we propose to remedy the off-policy issue by re-labeling the high-level transition $(s_t, g_t, \sum R_{t:t+c-1}, s_{t+c})$ with a different high-level action $\tilde{g}_t$ chosen to maximize the
probability $\mu^{lo}(a_{t:t+c-1}|s_{t:t+c-1}, \tilde{g}_{t:t+c-1})$, where the intermediate goals $\tilde{g}_{t+1:t+c-1}$ are computed using the fixed goal transition function $h$. In effect, each time we modify the low-level policy $\mu^{lo}$, we would like to answer the question: for which goals would this new controller have taken the same actions as the old one?

Most RL algorithms will use random action-space exploration to select actions, which means that the behavior policy (even for deterministic algorithms such as DDPG~\cite{ddpg}) is stochastic and the log probability $\log \mu^{lo}(a_{t:t+c-1}|s_{t:t+c-1}, \tilde{g}_{t:t+c-1})$ may be computed as
\begin{equation}
\log \mu^{lo}(a_{t:t+c-1}|s_{t:t+c-1}, \tilde{g}_{t:t+c-1}) \propto -\frac{1}{2}\sum_{i=t}^{t+c-1} ||a_i - \mu^{lo}(s_i, \tilde{g}_i)||_2^2 + \text{const}.
\label{eq:log_prob}
\end{equation}
To approximately maximize this quantity in practice, we compute this log probability for a number of goals $\tilde{g}_t$, and choose the maximal goal to re-label the experience.  
In our implementation, we calculate the quantity on eight candidate goals sampled randomly from a Gaussian centered at $s_{t+c} - s_t$.  We also include the original goal $g_t$ and a goal corresponding to the difference $s_{t+c} - s_t$ in the candidate set, to have a total of 10 candidates.  This provides a suitably diverse set of $\tilde{g}_t$ to approximately solve the $\arg\max$ of Equation~\ref{eq:log_prob}, while also biasing the result to be closer to candidates $\tilde{g}_t$ which we believe to be appropriate given our knowledge of the problem (see additional implementation details in the Appendix). 
%While more sophisticated optimization procedures could further improve the quality of the relabeled goal $\tilde{g}_i$, this simple procedure worked well in practice and has the appealing property of being very efficient and easy to parallelize with modern computation graph libraries.
Our approach here is only an approximation, and we elaborate on possible alternative off-policy corrections in the Appendix.

\section{Related Work}

Discovering meaningful and effective hierarchies of policies is a long standing research problem in RL~\cite{dayan1993feudal,parr1998reinforcement,sutton1999between,dietterich2000hierarchical,bacon2017option}. Classically, the work on HRL focused on discrete state domains, where state visitation and transition statistics can be used to construct heuristic sub-goals for low-level policies~\cite{stolle2002learning,mannor2004dynamic,chentanez2005intrinsically}. % and they can often be seen as learning state abstractions. 
The options framework~\cite{sutton1999between,precup2000temporal}, a popular formulation for HRL, proposes a termination policy for each sub-policy (option). 
While the traditional options framework relies on prior knowledge for designing options,% and primarily learned the high-level policy-over-options,
~\cite{bacon2017option} recently derived an actor-critic algorithm for learning them jointly with the higher-level policy. This option-critic architecture~\cite{bacon2017option} is an important step toward end-to-end HRL; however, such approaches are often prone to learning either a sub-policy that terminates every time step, or one effective sub-policy that runs through the whole episode. 
In practice, regularizers are essential to %encourage learning 
learn multiple effective and temporally abstracted 
sub-policies~\cite{bacon2017option,harb2017waiting,vezhnevets2016strategic}. %It is even more difficult to guarantee that these generalize and are semantically meaningful because they solely rely on the task rewards, which may be sparse.

To guarantee learning useful sub-policies, recent work has studied approaches that provide auxiliary rewards for the low-level policies~\cite{chentanez2005intrinsically,heess2016learning,kulkarni2016hierarchical,tessler2017deep,florensa2017stochastic}. These approaches rely on hand-crafted rewards based on prior domain knowledge~\cite{konidaris2007building,heess2016learning,kulkarni2016hierarchical,tessler2017deep} or  diversity-encouraging rewards like mutual information~\cite{daniel2012hierarchical,florensa2017stochastic}. 
A number of works have suggested that semantically distinct behavior can be induced by training on a set of diverse tasks, and have suggested pre-training the lower-level policy on such tasks~\cite{heess2016learning,florensa2017stochastic}, or training the multi-level hierarchical policy in a multi-task setup~\cite{frans2017meta,sigaud2018policy}.  However, having access to a collection of suitably similar tasks is a luxury which is not always available and may require hand-design.
%Most recently, an approach uses meta learning to discover a meaningful shared hierarchy through solving multiple environments~\cite{frans2017meta}. While this approach does not require explicitly hand-crafting a reward for each sub-policy, it still relies on designing a specific set of tasks such that solving them will encourage learning reusable policies for downstream tasks. 
Our method uses a generic reward that is specified with respect to the state space, and therefore avoids designing various rewards or multiple tasks.

Another difference from most HRL work~\cite{florensa2017stochastic,frans2017meta} is that we use off-policy learning, leading to significant improvements in sample efficiency. In end-to-end HRL, off-policy RL creates a non-stationary problem for the higher-level policy, since the lower-level is constantly changing. 
We are aware of only one recent work which applies HRL in an off-policy setting~\cite{levy2017hierarchical}.  As in our work, the authors devise a hierarchical structure in which a lower-level policy is trained to reach observations directed by a higher-level policy.  The multiple layers of policies are trained jointly in an off-policy manner, while ignoring the non-stationarity problem which we realize is a key issue for off-policy HRL.  
Accordingly, we derive and test an off-policy correction in the context of HRL, and empirically show that this technique is crucial to successfully train hierarchical policies on complex tasks.

Our work is related to FeUdal Networks (FuN)~\cite{vezhnevets2017feudal}, originally inspired from feudal RL~\cite{dayan1993feudal}. FuN also makes use of goals and a parameterized lower-level reward.  Unlike our method, FuN represents the goals and computes the rewards in terms of a learned state representation. In our experiments, we found this technique to under-perform compared to our approach, which uses the state in its raw form. We find that this has a number of benefits. For one, the lower-level policies can immediately begin receiving intrinsic rewards for reaching goals even before the higher-level policy receives a meaningful supervision signal from the task reward. Additionally, the representation is generic and simple to obtain. 
Goal-conditioned value functions~\cite{mahadevan2007proto,sutton2011horde,uvf,andrychowicz2017hindsight,pong2018temporal} are actively explored outside the context of HRL. Continued progress in this field may be used to further improve HRL methods. 

\section{Experiments}
\label{sec:experiments}

In our experiments, we compare HIRO method to prior techniques, and ablate the various components to understand their importance. Our experiments are conducted on a set of challenging environments that require a combination of locomotion and object manipulation. Visualizations of these environments are shown in Figure~\ref{fig:envs}.
See the Appendix for more details on each environment.

\textbf{Ant Gather.}
The ant gather task is a standard task introduced in~\cite{duan2016benchmarking}.  A simulated ant must navigate to gather apples while avoiding bombs, which are randomly placed in the environment at the beginning of each episode.
%In addition to observing its own position and velocity, the ant also observes depth readings of apples and bombs within its sensor range.
The ant receives a reward of $1$ for each apple and a reward of $-1$ for each bomb.

\textbf{Ant Maze.}
For the first difficult navigation task we adapted the maze environment introduced in~\cite{duan2016benchmarking}.
In this environment an ant must navigate to various locations in a `$\supset$'-shaped corridor.  We increase the default size of the maze so that the corridor is of width $8$.  
%The reward provided at each step is negative L2 distance to the target location.
In our evaluation, we assess the success rate of the policy when attempting to reach the end of the maze.

\textbf{Ant Push.}
In this task we introduce a movable block which the agent can interact with.  
%The reward at each step is negative L2 distance to the target.  
A greedy agent would move forward, unknowingly pushing the movable block until it blocks its path to the target.
To successfully reach the target, the ant must first move to the left around the block and then push the block right, clearing the path towards the target location.

\textbf{Ant Fall.}
This task extends the navigation to three dimensions.
The ant is placed on a raised platform, with the target location directly in front of it but separated by a chasm which it cannot traverse by itself.  Luckily, a movable block is provided on its right.  To successfully reach the target, the ant must first walk to the right, push the block into the chasm, and then safely cross.

\subsection{Comparative Analysis}

The primary comparisons to previous HRL methods are done with respect to 
%the option-critic architecture~\cite{bacon2017option}, 
FeUdal Networks (FuN)~\cite{vezhnevets2017feudal}, stochastic neural networks for HRL (SNN4HRL)~\cite{florensa2017stochastic}, and VIME~\cite{houthooft2016vime} (see Table~\ref{tab:comparisons}, and Appendix for more details).
As these algorithms often come with problem-specific design choices, we modify each for fairer comparisons. In terms of problem assumptions, our work is closest to that of FuN which is applicable to any single task without specific sub-policy reward engineering. MLSH~\cite{frans2017meta} is another promising recent work for HRL; however, since it relies on learning meaningful sub-policies through experiencing multiple, diverse, hand-designed tasks, we do not include explicit comparisons.  We leave exploring our method in the context of multi-task learning for future work.

%\textbf{Option-critic architecture}. The original option-critic architecture~\cite{bacon2017option} is designed for discrete action domains, so we modified the open-source code with DDPG-like amortization over $\arg\max_a Q(s,a,i)$. ... 
%{\bf TODO: Shane}

\begin{table}[t]
    \begin{center}
    \begin{tabular}{ c | c | c | c | c |}
    %\begin{tabular}{c c }
        %\multicolumn{2}{c}{\bf Average Reward}\\
        & {\bf Ant Gather} & {\bf Ant Maze} &
        {\bf Ant Push} & {\bf Ant Fall} \\
        \hline
        HIRO & {\bf 3.02}$\pm${\bf 1.49} & {\bf 0.99}$\pm${\bf 0.01} & {\bf 0.92}$\pm${\bf 0.04} & {\bf 0.66}$\pm${\bf 0.07} \\
        FuN representation & $0.03\pm 0.01$ & $0.0\pm 0.0$ & $0.0\pm 0.0$ &  $0.0\pm 0.0$ \\
        %\hline
        FuN transition PG & $0.41\pm 0.06$ & $0.0\pm 0.0$ & $0.56\pm 0.39$ & $0.01\pm 0.02$ \\
        FuN cos similarity & $0.85\pm 1.17$ & $0.16\pm 0.33$ & $0.06\pm 0.17$ & $0.07\pm 0.22$ \\
        FuN & $0.01\pm 0.01$ & $0.0\pm 0.0$ & $0.0\pm 0.0$ & $0.0\pm 0.0$ \\
        SNN4HRL & $1.92\pm 0.52$ & $0.0\pm 0.0$ & $0.02\pm 0.01$ & $0.0\pm 0.0$ \\
        VIME & $1.42\pm 0.90$ & $0.0\pm 0.0$ & $0.02\pm 0.02$ & $0.0\pm 0.0$ \\
        \hline
    \end{tabular}
    \end{center}
\caption{Performance of the best policy obtained in 10M steps of training, averaged over 10 randomly seeded trials with standard error.  Comparisons are to variants of FuN~\cite{vezhnevets2017feudal}, SNN4HRL~\cite{florensa2017stochastic}, and VIME~\cite{houthooft2016vime}. Even after extensive hyper-parameter searches, we were unable to achieve competitive performance from the baselines on any of our tasks.  In the Appendix, we include the only competitive result we could achieve -- VIME on Ant Gather trained for a much longer amount of time.
}
\label{tab:comparisons}
\end{table}

\begin{figure}
  \begin{center}
  	\begin{tabular}{cccc}
    \tiny Ant Gather & \tiny Ant Maze & \tiny Ant Push & \tiny Ant Fall \\
    \includegraphics[width=0.2\textwidth]{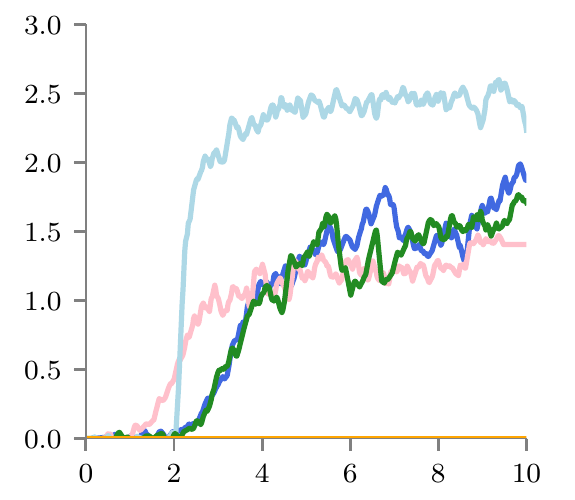} &
    \includegraphics[width=0.2\textwidth]{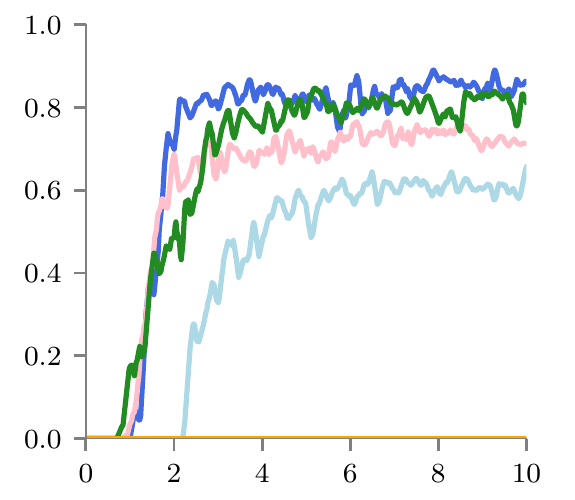} &
    \includegraphics[width=0.2\textwidth]{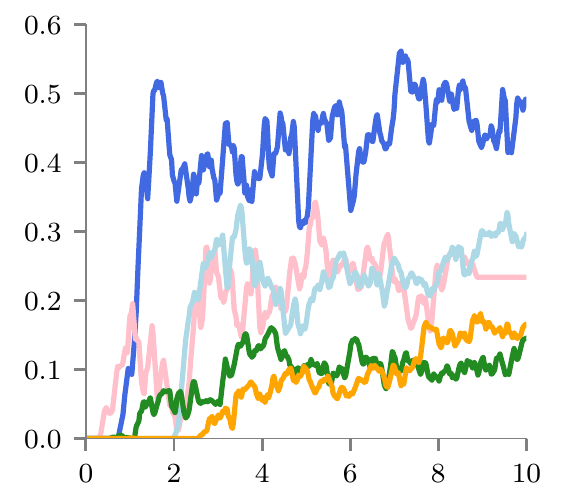} &
    \includegraphics[width=0.2\textwidth]{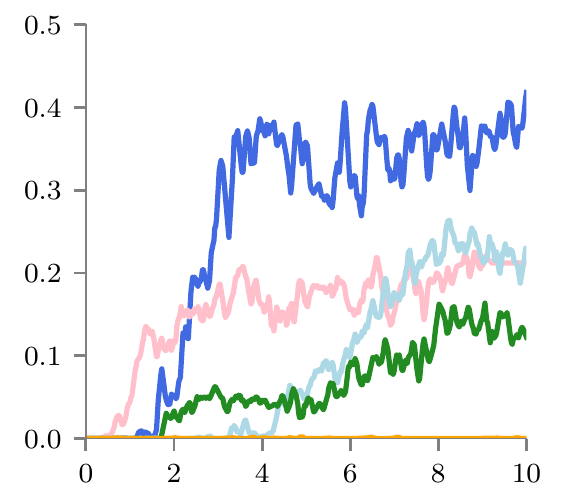} \\
    \multicolumn{4}{c}{\includegraphics[width=0.7\columnwidth]{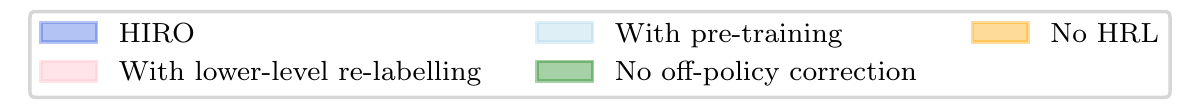}}
    \end{tabular}
  \end{center}
  \caption{Results of our method and a number of variants on a set of difficult tasks.  Each plot shows average reward (for Ant Gather) or average success rate (for the rest; see Appendix) over 10 randomly seeded trials, with x-axis in millions of environment steps.  We find that HIRO can perform well across all tasks.  We also note that HIRO learns rapidly; on the complex navigation tasks it requires only a few million environment steps (a few days in real-world interaction time) to achieve good performance.  Our method is only out-performed on Ant Gather by a variant that pre-trains the lower-level policy (thus not needing an off-policy correction).  
  %This suggests that there remain potential benefits to further improving the form of off-policy correction.
  }
	\label{fig:main_results}
\end{figure}

\textbf{FeUdal Network (FuN)}. Unlike SNN4HRL or VIME, the official open-source code for FuN was not available at the time of submission, and therefore we aimed to replicate key design choices of FuN from our algorithm implementation. FuN~\cite{vezhnevets2017feudal} primarily proposes four components: (1) transition policy gradient, (2) directional cosine similarity rewards, (3) goals specified with respect to a learned representation, and (4) dilated RNN. Since our tasks are low-dimensional and fully observed, we do not include design choice (4). 
For each of (1), (2), and (3), we apply an equivalent modification of our HRL method and evaluate its performance on the same tasks.
We also evaluate all modifications together as an approximation to the entire FuN paradigm.
Results in Table~\ref{tab:comparisons} show that on our tasks, the FuN modifications do not learn well, and other than Ant Gather are significantly out-performed by HIRO. In particular, it is worth noting that the use of learned representations, rather than observation goals, leads to almost no improvement on the tasks. This suggests that the choice of using goal observations as lower-level goals significantly improves HRL performance, by providing a strong supervision signal to the lower-level policy right from the beginning of training.

\textbf{Stochastic Neural Networks for HRL (SNN4HRL)}. SNN4HRL~\cite{florensa2017stochastic}
initially trains the low-level policy with a proxy reward to encourage learning useful diverse exploration policies, and then the high-level policy is trained in the tasks of interest while the low-level is fixed.  While SNN4HRL can perform better than FuN, it is still far behind our proposed HRL method.

\textbf{Variational Information Maximizing Exploration (VIME)}. VIME~\cite{houthooft2016vime} is not an HRL method but is used as a strong baseline in SNN4HRL. As discussed in~\cite{florensa2017stochastic} and matched by our results, for the benchmark's short horizon task of length 500, it performs approximately the same as SNN4HRL.

\textbf{Option-Critic Architecture}. We extended the option-critic architecture implementation~\cite{bacon2017option} for continuous actions and attempted a number of alternative variants besides the na\"{i}ve modification of the original. No versions yielded reasonable performance in our tasks, and so we omit it from the results. This is possibly due to difficulty in continuous control tasks, but most importantly the option-critic sub-policies rely solely on the external reward, making learning gait policies difficult.

\subsection{Ablative Analysis}
\label{sec:results}

In Figure~\ref{fig:main_results} we present results of our proposed HRL method (``HIRO'') compared with a number of variants to understand the importance of various design choices:

\textbf{With lower-level re-labelling.}  We evaluate the benefit of recent proposals~\cite{andrychowicz2017hindsight,tdm} to increase the amount of data available to an agent trained using a parameterized reward (the lower-level policy in our setup) by re-labeling experiences with randomly sampled goals.  This allows the lower-level policy to use experience collected with respect to a specific goal $g$ to be used to learn behavior with respect to any alternative goal $\tilde{g}$.  Our results show that this technique can provide an initial speed-up in training; however, its performance is quick to plateau. We hypothesize that re-labeling goals randomly may make lower-level training more difficult, since the policy must learn to not only satisfy the goals provided by the higher-level agent, but instead almost any conceivable goal. The benefit of re-labeling goals will require more research, and we encourage future work to investigate better ways to harness its benefits.

\textbf{With pre-training.}  In this variant we evaluate a simpler method to avoid the non-stationary issue in higher-level off-policy training.  Rather than correct for past experiences, we instead pre-train the lower-level policy for 2M steps (using goals sampled from a Gaussian) before freezing it and training the higher-level policy alone (this variant also has the advantage of allowing the higher-level policy to learn with respect to a deterministic, non-exploratory lower-level policy).  In the harder navigation tasks, we find that pre-training is detrimental.  This is understandable, as these tasks require specialization in different low-level behavior for different stages of the navigation.  By allowing the lower-level policy to continually learn as new parts of the environment are encountered, we are able to learn a lower-level policy which is better able to satisfy the desired goals of the higher-level.  In contrast, in the simpler and mostly homogeneous Ant Gather task, the advantage of pre-training is significant.  This suggests that our off-policy correction is still not perfect, and there is potentially significant benefit to be obtained by improving it.

\textbf{No off-policy correction.} We assess the advantage of including the off-policy correction compared to training off-policy na{\"i}vely, ignoring the non-stationary issue.  Interestingly, training an HRL policy this way can do quite well.  However, in the harder tasks (Ant Push, Ant Fall) the issue becomes difficult to ignore.  Accordingly, we observe a significant benefit from using the off-policy correction.

\textbf{No HRL.} Finally, we evaluate the ability of a single non-HRL policy to learn in these environments.  This variant makes almost no progress on the tasks compared to our HRL method.

\section{Conclusion}
We have presented a method for training a two-layer hierarchical policy.  Our approach is general, using learned goals to pass instructions from the higher-level policy to the lower-level one.  Moreover, we have described a method by which both polices may be trained in an off-policy manner concurrently for highly sample-efficient learning. Our experiments show that our method outperforms prior HRL algorithms and can solve exceedingly complex tasks that combine locomotion and rudimentary object interaction.
We note that our results are still far from perfect, and there is much work left for future research to improve the stability and performance of HRL methods on these tasks.

\section{Acknowledgments}
We thank Ben Eysenbach and others on the Google Brain
team for insightful comments and discussions.

\bibliography{ref}

\begin{thebibliography}{10}

\bibitem{andrychowicz2017hindsight}
Marcin Andrychowicz, Filip Wolski, Alex Ray, Jonas Schneider, Rachel Fong,
  Peter Welinder, Bob McGrew, Josh Tobin, OpenAI~Pieter Abbeel, and Wojciech
  Zaremba.
\newblock Hindsight experience replay.
\newblock In {\em Advances in Neural Information Processing Systems}, pages
  5048--5058, 2017.

\bibitem{bacon2017option}
Pierre-Luc Bacon, Jean Harb, and Doina Precup.
\newblock The option-critic architecture.
\newblock In {\em AAAI}, pages 1726--1734, 2017.

\bibitem{barth2018distributed}
Gabriel Barth-Maron, Matthew~W Hoffman, David Budden, Will Dabney, Dan Horgan,
  Alistair Muldal, Nicolas Heess, and Timothy Lillicrap.
\newblock Distributed distributional deterministic policy gradients.
\newblock {\em arXiv preprint arXiv:1804.08617}, 2018.

\bibitem{barto2003recent}
Andrew~G Barto and Sridhar Mahadevan.
\newblock Recent advances in hierarchical reinforcement learning.
\newblock {\em Discrete Event Dynamic Systems}, 13(4):341--379, 2003.

\bibitem{chentanez2005intrinsically}
Nuttapong Chentanez, Andrew~G Barto, and Satinder~P Singh.
\newblock Intrinsically motivated reinforcement learning.
\newblock In {\em Advances in neural information processing systems}, pages
  1281--1288, 2005.

\bibitem{daniel2012hierarchical}
Christian Daniel, Gerhard Neumann, and Jan Peters.
\newblock Hierarchical relative entropy policy search.
\newblock In {\em Artificial Intelligence and Statistics}, pages 273--281,
  2012.

\bibitem{dayan1993feudal}
Peter Dayan and Geoffrey~E Hinton.
\newblock Feudal reinforcement learning.
\newblock In {\em Advances in neural information processing systems}, pages
  271--278, 1993.

\bibitem{dietterich2000hierarchical}
Thomas~G Dietterich.
\newblock Hierarchical reinforcement learning with the maxq value function
  decomposition.
\newblock {\em Journal of Artificial Intelligence Research}, 13:227--303, 2000.

\bibitem{duan2016benchmarking}
Yan Duan, Xi~Chen, Rein Houthooft, John Schulman, and Pieter Abbeel.
\newblock Benchmarking deep reinforcement learning for continuous control.
\newblock In {\em International Conference on Machine Learning}, pages
  1329--1338, 2016.

\bibitem{florensa2017stochastic}
Carlos Florensa, Yan Duan, and Pieter Abbeel.
\newblock Stochastic neural networks for hierarchical reinforcement learning.
\newblock {\em arXiv preprint arXiv:1704.03012}, 2017.

\bibitem{frans2017meta}
Kevin Frans, Jonathan Ho, Xi~Chen, Pieter Abbeel, and John Schulman.
\newblock Meta learning shared hierarchies.
\newblock {\em International Conference on Learning Representations (ICLR)},
  2018.

\bibitem{td3}
Scott Fujimoto, Herke van Hoof, and Dave Meger.
\newblock Addressing function approximation error in actor-critic methods.
\newblock {\em arXiv preprint arXiv:1802.09477}, 2018.

\bibitem{gu2017deep}
Shixiang Gu, Ethan Holly, Timothy Lillicrap, and Sergey Levine.
\newblock Deep reinforcement learning for robotic manipulation with
  asynchronous off-policy updates.
\newblock In {\em Robotics and Automation (ICRA), 2017 IEEE International
  Conference on}, pages 3389--3396. IEEE, 2017.

\bibitem{gu2017interpolated}
Shixiang Gu, Tim Lillicrap, Richard~E Turner, Zoubin Ghahramani, Bernhard
  Sch{\"o}lkopf, and Sergey Levine.
\newblock Interpolated policy gradient: Merging on-policy and off-policy
  gradient estimation for deep reinforcement learning.
\newblock In {\em Advances in Neural Information Processing Systems}, pages
  3849--3858, 2017.

\bibitem{gu2016q}
Shixiang Gu, Timothy Lillicrap, Zoubin Ghahramani, Richard~E Turner, and Sergey
  Levine.
\newblock Q-prop: Sample-efficient policy gradient with an off-policy critic.
\newblock {\em arXiv preprint arXiv:1611.02247}, 2016.

\bibitem{sac}
Tuomas Haarnoja, Aurick Zhou, Pieter Abbeel, and Sergey Levine.
\newblock Soft actor-critic: Off-policy maximum entropy deep reinforcement
  learning with a stochastic actor.
\newblock {\em arXiv preprint arXiv:1801.01290}, 2018.

\bibitem{harb2017waiting}
Jean Harb, Pierre-Luc Bacon, Martin Klissarov, and Doina Precup.
\newblock When waiting is not an option: Learning options with a deliberation
  cost.
\newblock {\em arXiv preprint arXiv:1709.04571}, 2017.

\bibitem{heess2017emergence}
Nicolas Heess, Srinivasan Sriram, Jay Lemmon, Josh Merel, Greg Wayne, Yuval
  Tassa, Tom Erez, Ziyu Wang, Ali Eslami, Martin Riedmiller, et~al.
\newblock Emergence of locomotion behaviours in rich environments.
\newblock {\em arXiv preprint arXiv:1707.02286}, 2017.

\bibitem{heess2016learning}
Nicolas Heess, Greg Wayne, Yuval Tassa, Timothy Lillicrap, Martin Riedmiller,
  and David Silver.
\newblock Learning and transfer of modulated locomotor controllers.
\newblock {\em arXiv preprint arXiv:1610.05182}, 2016.

\bibitem{held2017automatic}
David Held, Xinyang Geng, Carlos Florensa, and Pieter Abbeel.
\newblock Automatic goal generation for reinforcement learning agents.
\newblock {\em arXiv preprint arXiv:1705.06366}, 2017.

\bibitem{houthooft2016vime}
Rein Houthooft, Xi~Chen, Yan Duan, John Schulman, Filip De~Turck, and Pieter
  Abbeel.
\newblock Vime: Variational information maximizing exploration.
\newblock In {\em Advances in Neural Information Processing Systems}, pages
  1109--1117, 2016.

\bibitem{kingma2013auto}
Diederik~P Kingma and Max Welling.
\newblock Auto-encoding variational bayes.
\newblock {\em arXiv preprint arXiv:1312.6114}, 2013.

\bibitem{konidaris2007building}
George Konidaris and Andrew~G Barto.
\newblock Building portable options: Skill transfer in reinforcement learning.
\newblock In {\em IJCAI}, volume~7, pages 895--900, 2007.

\bibitem{kulkarni2016hierarchical}
Tejas~D Kulkarni, Karthik Narasimhan, Ardavan Saeedi, and Josh Tenenbaum.
\newblock Hierarchical deep reinforcement learning: Integrating temporal
  abstraction and intrinsic motivation.
\newblock In {\em Advances in neural information processing systems}, pages
  3675--3683, 2016.

\bibitem{tdm}
Sergey Levine, Shane Gu, and Vitchyr Pong.
\newblock Temporal difference model learning: Model-free deep rl for
  model-based control.
\newblock 2018.

\bibitem{levy2017hierarchical}
Andrew Levy, Robert Platt, and Kate Saenko.
\newblock Hierarchical actor-critic.
\newblock {\em arXiv preprint arXiv:1712.00948}, 2017.

\bibitem{ddpg}
Timothy~P Lillicrap, Jonathan~J Hunt, Alexander Pritzel, Nicolas Heess, Tom
  Erez, Yuval Tassa, David Silver, and Daan Wierstra.
\newblock Continuous control with deep reinforcement learning.
\newblock {\em arXiv preprint arXiv:1509.02971}, 2015.

\bibitem{mahadevan2007proto}
Sridhar Mahadevan and Mauro Maggioni.
\newblock Proto-value functions: A laplacian framework for learning
  representation and control in markov decision processes.
\newblock {\em Journal of Machine Learning Research}, 8(Oct):2169--2231, 2007.

\bibitem{mannor2004dynamic}
Shie Mannor, Ishai Menache, Amit Hoze, and Uri Klein.
\newblock Dynamic abstraction in reinforcement learning via clustering.
\newblock In {\em Proceedings of the twenty-first international conference on
  Machine learning}, page~71. ACM, 2004.

\bibitem{munos2016safe}
R{\'e}mi Munos, Tom Stepleton, Anna Harutyunyan, and Marc Bellemare.
\newblock Safe and efficient off-policy reinforcement learning.
\newblock In {\em Advances in Neural Information Processing Systems}, pages
  1054--1062, 2016.

\bibitem{tpcl}
Ofir Nachum, Mohammad Norouzi, Kelvin Xu, and Dale Schuurmans.
\newblock Trust-pcl: An off-policy trust region method for continuous control.
\newblock {\em arXiv preprint arXiv:1707.01891}, 2017.

\bibitem{parr1998reinforcement}
Ronald Parr and Stuart~J Russell.
\newblock Reinforcement learning with hierarchies of machines.
\newblock In {\em Advances in neural information processing systems}, pages
  1043--1049, 1998.

\bibitem{plappert2018multi}
Matthias Plappert, Marcin Andrychowicz, Alex Ray, Bob McGrew, Bowen Baker,
  Glenn Powell, Jonas Schneider, Josh Tobin, Maciek Chociej, Peter Welinder,
  et~al.
\newblock Multi-goal reinforcement learning: Challenging robotics environments
  and request for research.
\newblock {\em arXiv preprint arXiv:1802.09464}, 2018.

\bibitem{pong2018temporal}
Vitchyr Pong, Shixiang Gu, Murtaza Dalal, and Sergey Levine.
\newblock Temporal difference models: Model-free deep rl for model-based
  control.
\newblock {\em International Conference on Learning Representations}, 2018.

\bibitem{precup2000temporal}
Doina Precup.
\newblock {\em Temporal abstraction in reinforcement learning}.
\newblock University of Massachusetts Amherst, 2000.

\bibitem{rajeswaran}
Aravind Rajeswaran, Vikash Kumar, Abhishek Gupta, John Schulman, Emanuel
  Todorov, and Sergey Levine.
\newblock Learning complex dexterous manipulation with deep reinforcement
  learning and demonstrations.
\newblock {\em arXiv preprint arXiv:1709.10087}, 2017.

\bibitem{rajeswaran2017towards}
Aravind Rajeswaran, Kendall Lowrey, Emanuel~V Todorov, and Sham~M Kakade.
\newblock Towards generalization and simplicity in continuous control.
\newblock In {\em Advances in Neural Information Processing Systems}, pages
  6553--6564, 2017.

\bibitem{uvf}
Tom Schaul, Daniel Horgan, Karol Gregor, and David Silver.
\newblock Universal value function approximators.
\newblock In {\em International Conference on Machine Learning}, pages
  1312--1320, 2015.

\bibitem{schulman2015trust}
John Schulman, Sergey Levine, Pieter Abbeel, Michael Jordan, and Philipp
  Moritz.
\newblock Trust region policy optimization.
\newblock In {\em International Conference on Machine Learning}, pages
  1889--1897, 2015.

\bibitem{sigaud2018policy}
Olivier Sigaud and Freek Stulp.
\newblock Policy search in continuous action domains: an overview.
\newblock {\em arXiv preprint arXiv:1803.04706}, 2018.

\bibitem{stolle2002learning}
Martin Stolle and Doina Precup.
\newblock Learning options in reinforcement learning.
\newblock In {\em International Symposium on abstraction, reformulation, and
  approximation}, pages 212--223. Springer, 2002.

\bibitem{sutton2011horde}
Richard~S Sutton, Joseph Modayil, Michael Delp, Thomas Degris, Patrick~M
  Pilarski, Adam White, and Doina Precup.
\newblock Horde: A scalable real-time architecture for learning knowledge from
  unsupervised sensorimotor interaction.
\newblock In {\em The 10th International Conference on Autonomous Agents and
  Multiagent Systems-Volume 2}, pages 761--768. International Foundation for
  Autonomous Agents and Multiagent Systems, 2011.

\bibitem{sutton1999between}
Richard~S Sutton, Doina Precup, and Satinder Singh.
\newblock Between mdps and semi-mdps: A framework for temporal abstraction in
  reinforcement learning.
\newblock {\em Artificial intelligence}, 112(1-2):181--211, 1999.

\bibitem{tessler2017deep}
Chen Tessler, Shahar Givony, Tom Zahavy, Daniel~J Mankowitz, and Shie Mannor.
\newblock A deep hierarchical approach to lifelong learning in minecraft.
\newblock In {\em AAAI}, volume~3, page~6, 2017.

\bibitem{mujoco}
Emanuel Todorov, Tom Erez, and Yuval Tassa.
\newblock Mujoco: A physics engine for model-based control.
\newblock In {\em Intelligent Robots and Systems (IROS), 2012 IEEE/RSJ
  International Conference on}, pages 5026--5033. IEEE, 2012.

\bibitem{vevcerik2017leveraging}
Matej Ve{\v{c}}er{\'\i}k, Todd Hester, Jonathan Scholz, Fumin Wang, Olivier
  Pietquin, Bilal Piot, Nicolas Heess, Thomas Roth{\"o}rl, Thomas Lampe, and
  Martin Riedmiller.
\newblock Leveraging demonstrations for deep reinforcement learning on robotics
  problems with sparse rewards.
\newblock {\em arXiv preprint arXiv:1707.08817}, 2017.

\bibitem{vezhnevets2016strategic}
Alexander Vezhnevets, Volodymyr Mnih, Simon Osindero, Alex Graves, Oriol
  Vinyals, John Agapiou, et~al.
\newblock Strategic attentive writer for learning macro-actions.
\newblock In {\em Advances in neural information processing systems}, pages
  3486--3494, 2016.

\bibitem{vezhnevets2017feudal}
Alexander~Sasha Vezhnevets, Simon Osindero, Tom Schaul, Nicolas Heess, Max
  Jaderberg, David Silver, and Koray Kavukcuoglu.
\newblock Feudal networks for hierarchical reinforcement learning.
\newblock {\em arXiv preprint arXiv:1703.01161}, 2017.

\bibitem{wang2017sample}
Ziyu Wang, Victor Bapst, Nicolas Heess, Volodymyr Mnih, Remi Munos, Koray
  Kavukcuoglu, and Nando de~Freitas.
\newblock Sample efficient actor-critic with experience replay.
\newblock {\em International Conference on Learning Representations}, 2017.

\end{thebibliography}
\bibliographystyle{plain}

%\section*{Appendix}
\appendix

\section{Discussion on Alternative Off-Policy Corrections for High-Level Actions}

Through our experiments, we found that our proposed maximum likelihood-based action relabeling works well empirically; however, we also tried other variants of off-policy correction schemes. While none of the methods below worked as well as ours in the tested domains based on preliminary experiments, we summarize them below as a reference for further future work on off-policy correction for HRL. 

The experience replay stores $(s_{t:t+c}, a_{t:t+c-1}, g_{t:t+c-1}, R_{t:t+c-1},s_{t+c})$ sampled from following a low-level policy $a_i\sim\mu_\beta^{lo}(a_i|s_i, g_i)$. $a_i$ is low-level action and $g_i$ is high-level action (or goal for the low-level policy). 
We want to estimate the following objective for the current low-level policy $\mu^{lo}(a|s,g)$, where $Q^{hi}$ represents the target network,
\begin{align}
 L(\theta) &= \mathbb{E}_{\beta}\left[ \left( Q^{hi}_\theta(s_t,g_t)  - y_t\right)^2 \right] \\
 y_t &= \mathbb{E}_{\prod_{i=t}^{t+c-1} \mu^{lo}(a_i|s_i,g_i)p(s_{i+1}|s_t,a_i)}\left[R_{t:t+c-1} + \gamma \max_{g} Q^{hi} (s_{t+c},g)\right] \\
 &= \mathbb{E}_{\prod_{i=t}^{t+c-1} \mu^{lo}_\beta(a_i|s_i,g_i)p(s_{i+1}|s_t,a_i)}\left[w_t\cdot\left(R_{t:t+c-1} + \gamma \max_{g} Q^{hi} (s_{t+c},g)\right)\right]\\ \label{eq:oco_def}
w_t &= \prod_{i=t}^{t+c-1} \frac{\mu^{lo}(a_i|s_i,g_i)}{\mu_\beta^{lo}(a_i|s_i,g_i)}.
\end{align}
We remind the reader that $g_i$ is computed using a deterministic dynamics from $g_t$ using $g_{i+1}=h(s_t,g_t,s_{t+1})=s_i + g_i - s_{i+1}$ for $i=t,t+1,\dots,t+c-2$.

\textbf{Direct Importance Correction}.
A na\"{i}ve approach is to directly use the unbiased estimator based on importance weighting defined by the expectation in Eq.~\ref{eq:oco_def},
\begin{align}
 L(\theta) &= \mathbb{E}_{\beta}\left[ \left( Q_\theta^{hi}(s_t,g_t)  - \hat{y}_t\right)^2 \right] \\
 \hat{y}_t &= w_t\left(R_{t:t+c-1} + \gamma \max_{g} Q^{hi} (s_{t+c},g)\right)\\
w_t &= \prod_{i=t}^{t+c-1} \frac{\mu^{lo}(a_i|s_i,g_i)}{\mu_\beta^{lo}(a_i|s_i,g_i)}.
\end{align}
For the continuous action domains in our paper, we found this estimator, while unbiased, has very high variance, and does not work well in practice.

\textbf{Importance-Based Action Relabeling}.
Instead of computing the high-variance importance weight for the sample goal $g_t$, we may also try to find a new goal $\tilde{g}_t$ such that the importance weight is approximately 1. This leads to the action relabeling objective as used in our method,
\begin{align}
 L(\theta) &= \mathbb{E}_{\beta}\left[ \left( Q_\theta^{hi}(s_t,\tilde{g}_t)  - \hat{y}_t\right)^2 \right] \\ \label{eq:act_reb}
 \hat{y}_t &= R_{t:t+c-1} + \gamma \max_{g} Q^{hi} (s_{t+c},g),
\end{align}
where $\tilde{g}_t$ can be found by minimizing loss functions such as,
\begin{align}
 \tilde{g}_t& = \arg\min_{g_t}\left(1-\prod_{i=t}^{t+c-1} \frac{\mu^{lo}(a_i|s_i,g_i)}{\mu_\beta^{lo}(a_i|s_i,g_i)}\right)^2\\
 \tilde{g}_t &= \arg\min_{g_t} \left(\sum_{i=t}^{t+c-1} \log \mu^{lo}(a_i|s_i,g_i)-\log\mu_\beta^{lo}(a_i|s_i,g_i)\right)^2.\label{eq:imp_act_reb} 
\end{align}
Since there is no guarantee that $\tilde{g}_t$ exists to make the loss function go to 0, this estimator is still biased. However, we could expect that the bias may be reduced.

\textbf{Model-Based Relabeling}. What we need to ensure for off-policy correction is that $(s_{t:t+c-1},g_{t:t+c-1},s_{t+c})$ is consistent with the dynamics of MDP transition $p(s_{i+1}|s_i,a_i)$ and current low-level policy $\mu^{lo}(a_i|s_i,g_i)$. If we can approximate either the high-level forward dynamics $\tilde{s}_{t+c}=p^{hi}(\cdot|s_t,g_t)$ or the inverse model $\tilde{g}_{t}\sim p^{hi}_{inv}(\cdot|s_t,s_{i+c})$, then we may directly do model-based prediction to relabel for either $s_{t+c}$ or $g_t$. While the action relabeling TD objective is given as Eq.~\ref{eq:act_reb}, the state relabeling objective is given by,
\begin{align}
 L(\theta) &= \mathbb{E}_{\beta}\left[ \left( Q_\theta^{hi}(s_t,g_t)  - \hat{y}_t\right)^2 \right] \\ \label{eq:act_reb2}
 \hat{y}_t &= R_{t:t+c-1} + \gamma \max_{g} Q^{hi} (\tilde{s}_{t+c},g).
\end{align}
The question is how to get $p^{hi}$ or $p^{hi}_{inv}$. While we can fit parametric functions on samples of data, this is often as difficult as fully model-based approach. We may instead make use of that fact that  the low-level is trying to reach the given goal states. Assuming the low-level policy eventually gets to complete the given goals, we may use the following forms,
\begin{align}
p^{hi}(\tilde{s}_{t+c}|s_t,g_t) = \mathcal{N}(s_t+g_t,\Sigma)\\
p^{hi}_{inv}(\tilde{g}_{t}|s_t,s_{t+c}) = \mathcal{N}(s_{t+1}-s_t,\Sigma).
\end{align}
This resembles transition policy gradient in FuN~\cite{vezhnevets2017feudal},
where the high-level policy is trained by assuming the low-level approximately completes the assigned goals.
Empirically, we did not observe this outperformed our approach on the tested domains.

\section{Environment Details}
Environments use the MuJoCo simulator~\cite{mujoco} with $dt=0.02$ and frame skip set to $5$.

\subsection{Gather}
We use the Gather environment provided by Rllab with a simulated ant agent.
The ant
is equivalent to the standard Rllab Ant, except that its gear range is reduced from $(-150, 150)$ to $(-30, 30)$.
In addition to observing $qpos$, $qvel$, and the current time step $t$, the agent also observes depth readings as defined by the standard Gather environment.  
We set the activity range to $10$ and the sensor span to $2\pi$, which matches the settings in~\cite{florensa2017stochastic}.

Each episode is terminated either when the ant falls or at 500 steps.

The reward used is the default reward (number of apples minus number of bombs).

\subsection{Navigation}
We devise three navigation tasks to evaluate our method.
In each navigation task, we create an environment of
$8\times8\times8$ blocks, some movable and some with fixed position.    
We use the same ant agent used in Gather.
The agent observes $qpos$, $qvel$, the current time step $t$, and the target location.  Its actions correspond to torques applied to joints.  At the beginning of each episode, the environment samples a target position $(g_x, g_y)$ and the agent is provided a reward at each step corresponding to negative L2 distance from the target: $-\sqrt{(g_x - x)^2 + (g_y - y)^2}$.  In one of the navigation tasks (Falling), the L2 distance is measured with respect to 3 coordinates: $x, y$, and $z$.  Each episode is 500 steps long (i.e., the episode does not terminate when the ant falls).

We describe the specifics of each navigation task below.

\subsubsection{Maze}
In this task, immovable blocks are placed to confine the agent to a ``$\supset$''-shaped corridor.  That is, blocks are placed everywhere except at $(0,0), (8,0), (16,0), (16,8), (16,16), (8,16), (0,16)$.  The agent is initialized at position $(0,0)$.  At each episode, a target position is sampled uniformly at random from $g_x\sim[-4, 20],g_y\sim[-4,20]$.

At evaluation time, we evaluate the agent only on its ability to reach $(0, 16)$.  We define a ``success'' as being within an L2 distance of $5$ from the target on the ultimate step of the episode.

\subsubsection{Push}
In this task, immovable blocks are placed everywhere except at $(0,0), (-8,0), (-8,8), (0,8), (8,8), (16,8), (0,16)$.  A movable block is placed at $(0, 8)$.  The agent is initialized at position $(0,0)$.  At each episode, the target position is fixed to $(g_x, g_y) = (0, 19)$.  Therefore, the agent must first move to the left, the push the movable block to the right, and then navigate to the target unimpeded.

At evaluation time, we evaluate the agent on its ability to reach $(0, 19)$.  We define a ``success'' as being within an L2 distance of $5$ from the target on the ultimate step of the episode.

\subsubsection{Fall}
In this task, the agent is initialized on a platform of height $4$.  Immovable blocks are placed everywhere except at $(-8,0), (0,0), (-8,8), (0,8), (-8,16), (0,16), (-8,24), (0,24)$.  
The raised platform is absent in the region $[-4,12]\times[12,20]$.
A movable block is placed at $(8,8)$.
The agent is initialized at position $(0,0,4.5)$.  At each episode, the target position is fixed to $(g_x, g_y, g_z) = (0, 27, 4.5)$.  Therefore, to cross the chasm, the agent must first push the movable block into the chasm and walk on top of it before navigating to the target.

At evaluation time, we evaluate the agent on its ability to reach $(0, 27, 4.5)$.  We define a ``success'' as being within an L2 distance of $5$ from the target on the ultimate step of the episode.

\section{Implementation Details}

\subsection{Network Structure}
We use the same basic network structure as proposed by the TD3 algorithm~\cite{td3}, with the only difference being that we use layers of size $(300, 300)$ rather than $(400, 300)$.

The output of the lower-level actor network (activated by $\tanh$) is scaled to the range of the low-level actions, which is $\pm30$.

The output of the higher-level actor network is scaled to an approximated range of high-level actions: $\pm10$ for the desired relative $x,y$; $\pm0.5$ for the desired relative $z$; $\pm1$ for the desired relative torso orientations; and the remaining limb angle ranges are available from the ant.xml file.

\subsection{Training Parameters}
\begin{itemize}
\item Discount $\gamma=0.99$ for both controllers.
\item Adam optimizer; actor learning rate $0.0001$; critic learning rate $0.001$.
\item Soft update targets $\tau=0.005$ for both controllers.
\item Replay buffer of size 200,000 for both controllers.
\item Lower-level train step and target update performed every 1 environment step.
\item Higher-level train step and target update performed every 10 environment steps.
\item No gradient clipping.
\item Reward scaling of $1.0$ for lower-level; $0.1$ for higher-level.
\item Lower-level exploration is Gaussian noise with $\sigma=1.0$.
\item Higher-level exploration is Gaussian noise with $\sigma=1.0$.
\end{itemize}

\subsection{Off-Policy Correction}
Given a high-level experience transition $(s_{t:t+c-1}, g_{t:t+c-1}, a_{t:t+c-1}, R_{t:t+c-1}, s_{t+c})$, we select 10 candidate $\tilde{g}_t$ to maximize the log-probability of the lower-level actions.  One is taken to be the original $g_t$; another to be $s_{t+c} - s_t$; and the remaining eight are sampled randomly from a Gaussian centered at $s_{t+c} - s_t$ with standard deviation $0.5\times\frac{1}{2}[\text{high-level action range}]$ (and subsequently clipped to lie within the high-level action range).

\subsection{Evaluation}
Learned hierarchical policies are evaluated every 50,000 training steps by averaging performance over 50 random episodes.

\section{Benchmark Details}

\subsection{FuN}
FuN~\cite{vezhnevets2017feudal} primarily proposes four components: (1) transition policy gradient, (2) directional cosine similarity rewards, (3) goals specified with respect to a learned representation, and (4) dilated RNN. Since our tasks are low-dimensional and fully observed, we do not include design choice (4). 
For each of (1), (2), and (3), we apply an equivalent modification of our HRL method and evaluate its performance on the same tasks.
For representation learning, we augment our method with a two-hidden-layer feed-forward neural network for embedding the observations before passing them to the lower and higher-level policies.  The higher-level policy specifies high-level actions and rewards low-level behavior with respect to this representation.
For the transition policy gradient, we modify our off-policy correction to instead replace a goal $g_t$ with a goal $\tilde{g}_t$ sampled from a Gaussian centered at $s_{t+c} - s_t$, with standard deviation set to $0.1\times\frac{1}{2}[\text{high-level action range}]$.  This is analogous to FuN's transition policy gradient, which trains the higher-level policy under the assumption that its state transitions are distributed symmetrically around its proposed goals.
For directional rewards, we replace our relative position parameterized reward function $r$ with a cosine similarity reward function equivalent to that used in FuN.

\begin{figure}
  \begin{center}
  	\begin{tabular}{c}
    \tiny Ant Gather \\
    \includegraphics[width=0.25\textwidth]{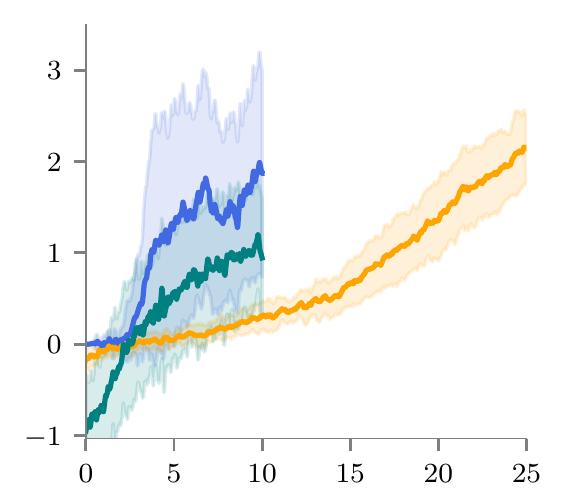} \\
    \multicolumn{1}{c}{\includegraphics[width=0.6\columnwidth]{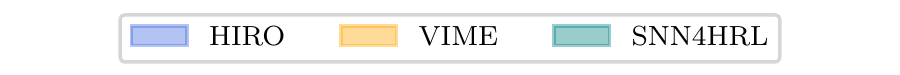}}
    \end{tabular}
  \end{center}
  \caption{Performance of HIRO compared to VIME and SNN4HRL, averaged over 10 trials with x-axis in millions of experience samples. After a hyper-parameter search on the baselines, we were only able to get competitive performance with HIRO from VIME on Ant Gather, with a significantly higher amount of experience.  On the other tasks, we were unable to achieve good baseline performance, even with more experience. The SNN4HRL curve does not include 25M transitions used in pre-training.
  }
	\label{fig:vime}
\end{figure}

\subsection{SNN4HRL}
SNN4HRL~\cite{florensa2017stochastic} trains the hierarchical policy stage-wise. It first uses a proxy reward with mutual information (MI) bonus to learn a mixture of low-level skills encouraging diverse movements, and then trains a high-level policy that controls switching among these skills to optimize for the task reward. We imported our additional environments to the official open-source codes with minimal modifications and followed the similar benchmark setup as in~\cite{florensa2017stochastic}. All policies are trained with TRPO~\cite{schulman2015trust} with step size 0.01 and discount 0.99. All neural networks (the SNN, the Latent Regressor Network and the Manager Network) have 2 layers of 32 hidden units as done in~\cite{florensa2017stochastic}\footnote{While the policy network sizes are significantly smaller than those used for our method, we observed no significant improvements with larger network sizes and this observation conforms with prior results that on-policy policy gradient methods can perform well on MuJoCo benchmark tasks with very small networks~\cite{duan2016benchmarking,rajeswaran2017towards}.}. We report the final results averaged over 10 random seeds, where we ran the low-level training using 5 seeds, and ran the high-level training using 2 seeds per trained low-level policy.

\begin{figure}
    \begin{center}
  	\begin{tabular}{ccc}
    \includegraphics[height=0.385\textwidth]{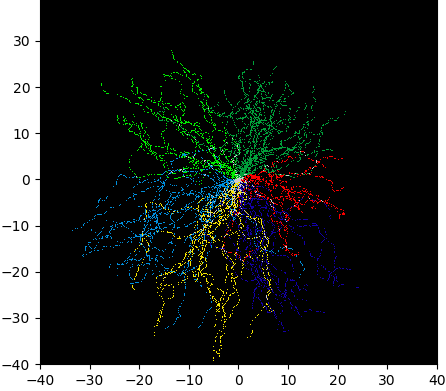} &
    \includegraphics[height=0.385\textwidth]{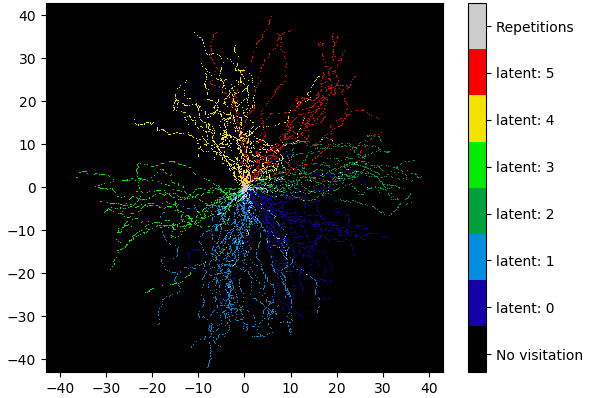}
    %\multicolumn{3}{c}{\includegraphics[width=0.4\textwidth]{ant_fall_b}}
    \end{tabular}
  \end{center}
  \caption{Visitation plots for 2 random seeds for the low-level SNN policy in the SNN4HRL benchmark. All 6 policies diversify in different exploration directions.}
	\label{fig:visitations}
\end{figure}

For the skill training, the mesh density used to grid the (x, y) space and give the MI bonus is 10 divisions/unit, and the mutual information bonus coefficient $\alpha_H$ is set to $10$. Bilinear integration is used for the SNN. The number of skills trained is 6. The batch size and the maximum path length for the pre-train task are 50,000 and 500 respectively. To guarantee learning effective skills, we trained the policy with 500 updates, or 25M transitions. The visualization of learned skill primitives is shown in Figure~\ref{fig:visitations}, where the learned 6 sub-policies successfully cover different exploration directions. 

For the high-level policy learning\footnote{In both SNN4HRL~\cite{florensa2017stochastic} and VIME~\cite{houthooft2016vime}, primarily the results are reported and compared on SwimmerMaze and SwimmerGather, and therefore the experimental results are different.}, we fixed the low-level policies as done in SNN4HRL and trained a policy that outputs a categorical action every 10 time steps that choose which skill to execute. For fair comparisons, we experimented with both sparse and dense rewards for the maze environments, and searched over batch sizes for $(1e4,5e4,5e5)$ transitions. We observed that the dense rewards did not help for SNN4HRL significantly, since the policy often quickly converge to local optimum. We found the batch size of $1e4$ is too noisy, and the batch size of $5e5$ is unnecessarily sample intensive, so the high-level policy is trained using batch size of $5e4$, the default value in their paper, for 300 updates, or 15M transitions. The combined training sample size of 40M is generously more than 10M used for our methods; however, our method still outperforms these SNN4HRL results substantially. 

\subsection{Variational Information Maximizing Exploration}
Variational Information Maximizing Exploration (VIME)~\cite{houthooft2016vime}, while not a HRL algorithm, exhibits good performance on prior benchmark maze and gather tasks, and is also used as a strong baseline in SNN4HRL~\cite{florensa2017stochastic}. We ran the algorithm using the default settings in the official open-source implementation. Batch size of 50,000 is used. We report the average performance across 5 seeds after running the algorithm for 300 updates, or 15M transitions. Only the Gather task required more samples to converge to the final performance, and required 25M+ transitions to reach the same performance as what our method reached in a few million transitions.

\subsection{Option-Critic Architecture}
We also experimented with continuous-action variants of the option-critic architecture~\cite{bacon2017option}. The option-policy $\pi_{\omega,\theta}(a|s)$ for option $\omega$ is parameterized as a Gaussian, whose mean is output from a neural network taking in $s$ and $\omega$, and variance is chosen to be global and diagonal. We first tested naively extending the official open-source implementation for continuous action, and then tried modifying the learning procedure such that the critic learns the state-option-action value function $Q_U(s,\omega,a)$ instead of the state-option value function $Q_\Omega(s,\omega)$ in the original implementation. This creates slight changes for the value and policy training objectives, while the loss for termination policy $\beta_{\omega,\nu}(s)$ is basically kept the same. Concretely, for the first variant, 
we trained $Q_\Omega(s,\omega)$ and the option-policy $\pi_{\omega,\theta}(a|s)$ with the following gradients,
\begin{align}
&g_{\Omega} = \mathbb{E}_{s_t,\omega_t,s_{t+1}\sim\beta}\left[\frac{\partial }{\partial \Omega}\left( Q_\Omega(s_t,\omega_t) - y_t \right)^2 \right] \\
&g_\theta = \mathbb{E}_{s_t,\omega_t,a_t,s_{t+1}\sim\pi} \left[ \left( y_t - b_t  \right)\nabla_\theta \log \pi_{\omega_t,\theta}(a_t|s_t) \right] \\
& \quad y_t = r_{t+1} + \gamma \left(\left(1-\beta_{\omega_t,\nu}(s_{t+1})\right)Q'(s_{t+1},\omega_t) + \beta_{\omega_t,\nu}(s_{t+1})\max_{\omega}Q'(s_{t+1},\omega)\right)
\end{align}
where $Q'$ represents the target network, and $\beta$ and $\pi$ represent using off-policy and on-policy transition samples respectively. For simplicity of explanation, we assumed that the reward only depends on states, but similar arguments can be made for the general case. There are two pragmatic problems for this objective. First, the policy gradient, which relies on a score function estimate, could be high variance especially with respect to a continuous policy $\pi_{\omega,\theta}$. We experimented with several choices of baselines $b_t$, including $Q_\Omega(s_t,\omega_t)$ and $Q;(s_t,\omega_t)$. The second problem is that the off-policy learning for $Q_\Omega(s_t,\omega_t)$ does not use the action $a_t$ taken and only relies on $\omega_t$. This effectively creates the same non-stationarity problem with respect to the high-level policy as our method, since it ignores that for the same $\omega_t$ and $s_t$, the next state $s_{t+1}$ can be different due to changing $\pi_{\omega,\theta}$. To counter both problems, we also explored another variant of the option-critic implementation at the expense of potentially more computation and network parameters, which conforms more closely with the policy gradient theorems in the original paper. 
Specifically,  
we trained $Q_U(s,\omega,a)$ and the option-policy $\pi_{\omega,\theta}(a|s)$ with the following gradients,
\begin{align}
&g_{U} = \mathbb{E}_{s_t,\omega_t,a_t,s_{t+1}\sim\beta}\left[\frac{\partial }{\partial U}\left( Q_U(s_t,\omega_t,a_t) - y_t \right)^2 \right] \\
&g_\theta = \mathbb{E}_{s_t,\omega_t\sim\pi} \left[ \nabla_\theta \mathbb{E}_{a\sim\pi_{\omega_t,\theta}(a|s_t)}\left[Q_U(s_t,\omega_t,a)\right] \right] \\
& \quad y_t = r_{t+1} + \gamma \left(\left(1-\beta_{\omega_t,\nu}(s_{t+1})\right)Q'(s_{t+1},\omega_t) + \beta_{\omega_t,\nu}(s_{t+1})\max_{\omega}Q'(s_{t+1},\omega)\right)\\
&\quad Q'(s,\omega) = \mathbb{E}_{a\sim\pi_{\omega,\theta}(a|s)}\left[Q'(s,\omega,a)\right].
\end{align}
In this implementation, we observe that the off-policy learning for $Q_U(s,\omega,a)$ can effectively utilize both $\omega_t$ and $a_t$, removing the non-stationarity problem, and the policy gradient can be estimated with lower variance using reparametrization trick~\cite{kingma2013auto} through the critic directly. Furthermore, since the policy gradient no longer requires next state estimate, off-policy state samples may also be used along with enumeration over all $\omega$,
\begin{align}
&g_\theta = \mathbb{E}_{s_t\sim\beta} \left[ \sum_\omega \nabla_\theta \mathbb{E}_{a\sim\pi_{\omega,\theta}(a|s_t)}\left[Q_U(s_t,\omega,a)\right] \right].
\end{align}
Making similar approximations for the termination policy, this enables a fully off-policy actor-critic algorithm like DDPG~\cite{ddpg} for the option-critic architecture.

While we tried these modifications, we could not make the option-critic implementation work reasonably on our domains. The main difficulty is likely because the low-level option-policies are learned using only the external task reward, a limitation in a direct end-to-end hierarchical policy structure. While in our experiments we could not show substantial successes, the algorithm may work better with more sophisticated modifications to the policy evaluation or policy improvement routines based on recent advances~\cite{munos2016safe,wang2017sample,gu2017interpolated,sac,td3}, and we leave further comparisons for future work.

\end{document}